\documentclass[a4paper]{article}

\usepackage{INTERSPEECH2021}
\usepackage{eqparbox}
\usepackage{microtype}
\usepackage{cite}
\usepackage{amsmath}
\usepackage{etoolbox,siunitx}
\robustify\bfseries
\usepackage[breaklinks=true,colorlinks,bookmarks=false]{hyperref}
\usepackage{enumitem}
\setlist[itemize]{wide=0pt, leftmargin=10pt, labelwidth=5pt, align=left}

\title{4-bit Quantization of LSTM-based Speech Recognition Models}
\name{Andrea Fasoli$^{1\dagger}$, Chia-Yu Chen$^{2\dagger}$, Mauricio Serrano$^2$, Xiao Sun$^2$, Naigang Wang$^2$,\\ Swagath Venkataramani$^1$, George Saon$^2$, Xiaodong Cui$^2$, Brian Kingsbury$^2$, Wei Zhang$^2$,\\ Zolt{\'a}n T{\"u}ske$^1$, Kailash Gopalakrishnan$^2$}
\address{
  IBM Research, USA\\
  $^\dagger$These authors contributed equally to this work}
\email{$^1$\{andrea.fasoli,swagath.venkataramani,zoltan.tuske\}@ibm.com\\
$^2$\{cchen,mserrano,xsun,nwang,gsaon,cuix,bedk,weiz,kailash\}@us.ibm.com
}

\begin{document}
\maketitle
\begin{abstract}

We investigate the impact of aggressive low-precision representations of weights and activations in two families of large LSTM-based architectures for Automatic Speech Recognition (ASR): hybrid Deep Bidirectional LSTM - Hidden Markov Models (DBLSTM-HMMs) and Recurrent Neural Network - Transducers (RNN-Ts). Using a 4-bit integer representation, a na\"ive quantization approach applied to the LSTM portion of these models results in significant Word Error Rate (WER) degradation. On the other hand, we show that minimal accuracy loss is achievable with an appropriate choice of quantizers and initializations. In particular, we customize quantization schemes depending on the local properties of the network, improving recognition performance while limiting computational time. We demonstrate our solution on the Switchboard (SWB) and CallHome (CH) test sets of the NIST Hub5-2000 evaluation. DBLSTM-HMMs trained with 300 or 2000~hours of SWB data achieves $<$0.5\% and $<$1\% average WER degradation, respectively. On the more challenging RNN-T models, our quantization strategy limits degradation in 4-bit inference to 1.3\%.

\end{abstract}

\section{Introduction} \label{sec:intro}

Neural Network (NN) models for ASR have achieved tremendous success in the past decade, significantly reducing the gap with human performance~\cite{saon2017english, tuske2020single}. Long Short-Term Memory (LSTM) building blocks are at the core of ASR models, either in NN-HMM hybrid~\cite{saon2017english} or NN end-to-end form~\cite{tuske2020single,saon2021advancing}, delivering state-of-the-art performance on standard speech benchmarks, such as the SWB corpus~\cite{godfrey1992switchboard}.

Many strategies have been proposed to alleviate the challenges posed by the large size and protracted training time of ASR models (these challenges get exacerbated during low-precision quantization emulation) - limiting their application in resource constrained scenarios. These include knowledge distillation~\cite{wong2016sequence, chebotar2016distilling}, low-rank matrix factorization~\cite{sainath2013low}, sparsification~\cite{seide2011conversational}, and quantization~\cite{hubara2017quantized}. These methods aim to reduce model size, computation, and time to convergence.
\begin{figure}[t]
  \centering
  \includegraphics[width=\linewidth]{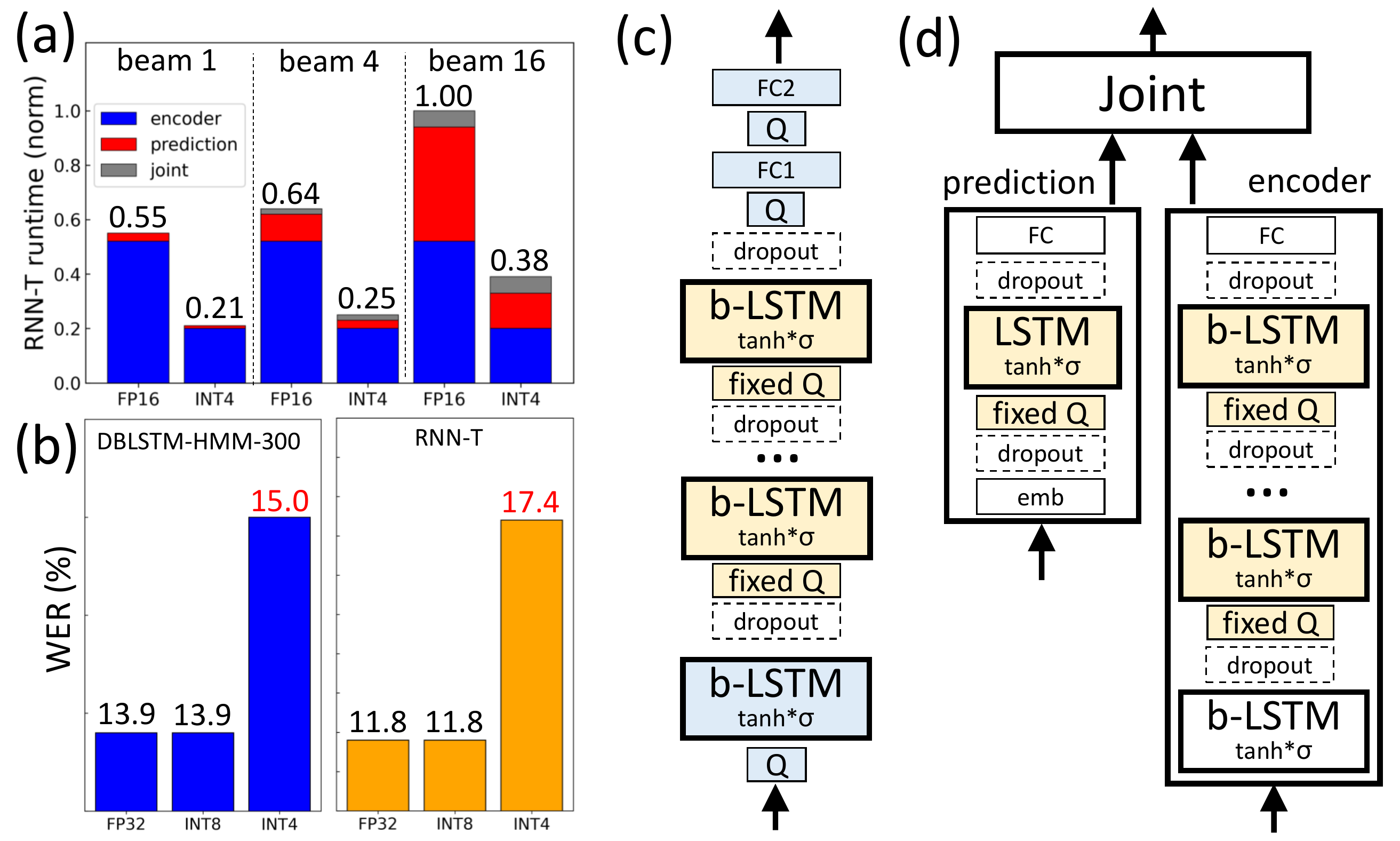}
  \caption{(a) Estimated RNN-T inference time. (b) Average WER for SWB/CH inference at different bit-widths. (c-d) Architecture of DBLSTM-HMM and RNN-T with learnable (blue) or fixed (yellow) quantization using BAC.}
  \label{fig:problem-intro}
\vspace{-5mm}
\end{figure}
Driven by advances in low-precision arithmetic and training algorithms, quantization is becoming a key methodology to speed up inference for ASR models~\cite{he2019streaming, shangguan2019optimizing}. In a quantized model, parameters are stored and computations are performed with lower bit widths than the full-precision counterpart, which is frequently IEEE \texttt{float32} (FP32). With Quantization Aware Training (QAT), the quantization error is introduced during the training phase via operations in low-precision arithmetic. This allows the network to progressively learn to incorporate the impact of reduced precision of its weights and activations and typically improves performance over Post-Training Quantization schemes~\cite{wu2020integer}.

Runtime estimates of 4-bit integer (INT4) quantized ASR models (Fig.~\ref{fig:problem-intro}(a)) show 2.6$\times$ faster streaming end-to-end speech recognition compared to 16-bit floating point (FP), a widespread format for model deployment nowadays (see Sec.~\ref{ssec:runtime} for more details). However, aggressive quantization of LSTM-based models can severely impact performance, especially in large models trained on extensive speech data sets. For example, Fig.~\ref{fig:problem-intro}(b) shows that quantization of weights and activations of a DBLSTM-HMM model with a simple quantizer results in no WER degradation for INT8, but in a $>$1\% degradation for INT4. Moreover, it is observed that INT4 quantization of RNN-T models shows $>$5\% degradation, while INT8 quantization shows the same accuracy as the FP32 model (Fig.~\ref{fig:problem-intro}(b)). Although multiple techniques have been proposed to mitigate the performance deterioration, they are limited to 8-bit quantization~\cite{he2019streaming, shangguan2019optimizing, wu2020integer, nguyen2020quantization} and to small LSTM networks (1 or 2 layers) on simple train and test datasets (e.g., the Penn Tree Bank (PTB))~\cite{hubara2017quantized, ardakani2018learning, hou2019normalization, he2016effective, xu2018alternating, xu2020low}, while viable 4-bit solutions for industrial scale models remain elusive.

Quantization of large LSTM-based speech models to 4 bits faces two primary challenges. (a) \textbf{Quantizer selection}: in INT4 quantization, simple quantizers fail to deliver performance comparable to full-precision models. Alternative strategies that learn quantization parameters during QAT are less effective than in non-ASR tasks due to the inherent constraints set by the LSTM cell. (b) \textbf{QAT training time}: conversion of FP32 values to lower precisions during QAT training can be costly and slow down training significantly, making the optimization of large networks on large datasets a particularly daunting task.

In this work, we address these challenges and propose a practical methodology to quantize both weights and activations in state-of-the-art speech models using an INT4 representation. Our contributions are as follow:
\vspace{-0.052cm}  
    \setlist{nolistsep}
    \begin{itemize}[noitemsep]
    \item We introduce a novel quantizer (Bound Aware Clipping, BAC) that achieves minimal degradation for 4-bit quantization while speeding up QAT in LSTM-based models. 
    \item We adopt QAT fine-tuning techniques with custom learning rate policies, addressing the training time bottleneck and shortening training duration by over 50\%. 
    \item Experimentally, we demonstrate record accuracy in large-scale 4-bit LSTM based models (DBLSTM-HMM and RNN-T) trained on large speech data sets (SWB 300~h and 2000~h).
\end{itemize}

\section{Speech models} \label{sec:models}

We investigate the impact of INT4 quantization on ASR performance of two families of LSTM-based architectures: Hybrid DBLSTM-HMMs and RNN-T. For all networks, large training data sets (SWB) are used and testing is done on the SWB and CH test sets (NIST Hub5-2000 evaluation).

\textbf{DBLSTM-HMM-300} (Fig.~\ref{fig:problem-intro}(c)): the network comprises 4 bi-directional LSTM layers (input size 140, hidden size 512 per direction, sequence length 21) followed by 2 fully connected (FC) layers in bottleneck configuration with 32k outputs corresponding to context-dependent hidden Markov model states \cite{cui2017embedding}. Decoding is performed with a 4M 4-gram language model. The training set consists of 262 h of SWB 1 audio with transcripts. The optimizer is SGD, with initial LR of 0.1, annealed by $\frac{1}{\sqrt{2}}$ from epoch 10. We train for 20 epochs on a V100 GPU with batch size 128.

\textbf{DBLSTM-HMM-2000} (Fig.~\ref{fig:problem-intro}(c)): the architecture is equivalent to DBLSTM-HMM-300 except for the use of 6 LSTM layers and input size 260. We train with 1975 h of audio: 262~h from SWB, 1698~h from the Fisher corpus, and 15~h of CH audio. We use a total batch size of 4096 (16 V100 GPUs) and the scaled LR schedule with a warm-up period in the first 10 epochs~\cite{zhang2019highly}.

\textbf{RNN-T} (Fig.~\ref{fig:problem-intro}(d)): the acoustic encoder consists of 6 bi-directional LSTM layers (input size 240 + 100-dim i-vectors, hidden size 640 per direction, variable sequence length). The prediction network is a single unidirectional LSTM layer with 768 cells. Encoder and prediction network outputs are combined multiplicatively in a joint network~\cite{saon2021advancing}, with an FC layer and log-Softmax over 46 output characters. We train for 20 epochs with batch size 64, using AdamW and a triangular LR policy (OneCycleLR), on the audio and character-level transcripts from the SWB corpus, augmented with speed and tempo perturbation~\cite{ko2015audio}, SpecAugment~\cite{park2019specaugment}, and Sequence Noise Injection~\cite{saon2019sequence}.
%We use a total batch size of 64 and train for 20 epochs, using AdamW and a triangular LR policy (OneCycleLR), on the audio and character-level transcripts from the SWB corpus, augmented with speed and tempo perturbation~\cite{ko2015audio}, SpecAugment~\cite{park2019specaugment}, and Sequence Noise Injection~\cite{saon2019sequence}.

\section{Quantization schemes} \label{sec:quantization_schemes}

\subsection{Quantizers} \label{ssec:quantizers}

Linear Quantization (LQ) is a widely used technique to approximate full-precision values. This is achieved by mapping them onto an interval discretized into $2^{\text{bitwidth}}$ levels. Various approaches exist to determine the boundaries of the interval (or, equivalently, the scaling factor applied to the full-precision values). LQ allows for both symmetric and asymmetric representations, with respect to zero.

Symmetric LQ takes the form:
\begin{equation}\label{eq:symmetric_lq}
    y_q = clamp\bigl( round( (y + \alpha) \cdot \frac{2^k-1}{2\alpha}) \cdot \frac{2\alpha}{2^k - 1} -\alpha, -\alpha, \alpha\bigr)
\end{equation}
with $y$ and $y_q$ being the full-precision and quantized tensor, respectively, $k$ the bit-width, and $\alpha$ the quantization boundary. The $clamp$ function limits the range of $y$ to [$-\alpha$,$\alpha$]. In \textit{asymmetric} LQ, two independent parameters $\alpha_+$ and $\alpha_-$ define the upper and lower boundary, respectively, and $y$ is scaled using $\frac{2^k-1}{\alpha_+ - \alpha_-}$.

The most common quantizer is symmetric and simply defines $\alpha = \max{|y|}$. Herein, we refer to this quantizer as MAX. MAX can be applied to both weight and activation quantization.

Alternatively, it was shown that knowledge of the weight distribution statistics allows for an improved representation in the low-precision space~\cite{choi2018bridging}. In this context, Statistics-Aware Weight Binning (SAWB), a symmetric LQ approach, aims to identify the clamp boundary $\alpha$ as an approximation of the optimal $\alpha^*$ that minimizes the mean square error introduced by the quantization process. During training, $\alpha$ can be calculated as the linear combination of the expected values $E(|y|)$ and $E(y^2)$, using empirical, bit-width-dependent coefficients.

With Parameterized Clipping Activations (PACT)~\cite{choi2018bridging}, the boundary $\alpha$ for activation quantization is learned. The asymmetric version of PACT uses two independently-learned boundaries, $\alpha_+$ and $\alpha_-$. The gradients $\frac{\partial L\hphantom{_\pm}}{\partial \alpha_{\pm}} = \frac{\partial L\hphantom{_q}}{\partial y_q} \frac{\partial y_q}{\partial \tilde{y}} \frac{\partial \tilde{y}\hphantom{_\pm}}{\partial \alpha_{\pm}}$, with $\tilde{y}$ being the clamped value, are derived via the Straight-Through Estimator (STE)~\cite{bengio2013estimating}, which estimates $\frac{\partial y_q}{\partial \tilde{y}}$ as 1.

In order to address the peculiar features of LSTM-based networks, we introduce a new quantizer, Bound Aware Clipping (BAC), which selectively applies different quantization strategies to different layers. Due to the large dynamic range of inputs fed into the \textit{first layer} of the network, it is beneficial to either learn the most appropriate $\alpha_{\pm}$ boundaries of the quantizer during training (as in PACT) or, alternatively, to process it in high precision. On the other hand, due to the inherently restricted nature of LSTM activations, fixed boundaries are a natural choice for layers beyond the first. The use of unchanging boundaries saves computational time, providing very relevant speed-ups in some networks. As the presence of dropout scales the inputs in-training outside the initial [-1,1] range defined by the $tanh\cdot\sigma$ operation within the LSTM cell, BAC sets symmetric boundaries for the inputs using $\alpha = \frac{1}{1 - dropout}$ (corresponding to the dropout scaling factor), and symmetric boundaries for the hidden states with $\alpha = 1$, as these are not modified by dropout.

An additional effect of using dropout to regularize LSTM networks is the large number of zeros it generates. Their poor representation due to misalignment of the discrete quantization levels can result in a very significant quantization error. For INT4 quantization, an effective mitigation strategy is to use $2^{bits}-1$ levels, such that, for symmetric quantizers, the middle interval always represents zeros exactly. We apply this strategy in quantizing activations of the models introduced in Sec.~\ref{sec:models}. 

\subsection{Model quantization} \label{ssec:models_quant}

In the DBLSTM-HMM models, we quantize weights (without biases) and activations of all LSTM layers (both directions) as well as the two FC layers. With reference to Fig.~\ref{fig:problem-intro}(c), we use our selective quantizer BAC, such that the first LSTM layer uses learnable parameters, as in PACT, while fixed boundaries are used in the subsequent LSTM layers. Full weight quantization results, for both models, in 99.8\% of the network parameters being quantized. Inputs and hidden states are quantized at every call of the LSTM cell, while cell states are preserved in full-precision. The rationale is that they only participate in element-wise multiplications, resulting in a very limited contribution ($\sim$0.1\%) to the total FLOPs. The HMM decoder is not quantized.

In the RNN-T (Fig.~\ref{fig:problem-intro}(d)), we quantize weights (without biases) and activations for the LSTM layers (both directions) in the acoustic encoder. The first layer, accounting for 8.8\% of the total parameters ($~$4.8\% of total inference computation with beam size 16), is either quantized or kept in full precision (see Sec.~\ref{ssec:rnnt}). BAC with fixed clip values is applied in the subsequent LSTM layers. The prediction network is also quantized: while carrying only 4.1\% of the total parameters, it is repeatedly executed at inference time as part of a beam search algorithm. Consequently, a quantized prediction network speeds up inference time by 2.6$\times$, compared to a FP16 RNN-T (see Sec.~\ref{ssec:runtime}). FC layers and the joint network account for $<$1\% of total parameters and are kept in full-precision. With this configuration, 90.2\% of the parameters are INT4.

\begin{figure}[t]
  \centering
  \includegraphics[width=\linewidth]{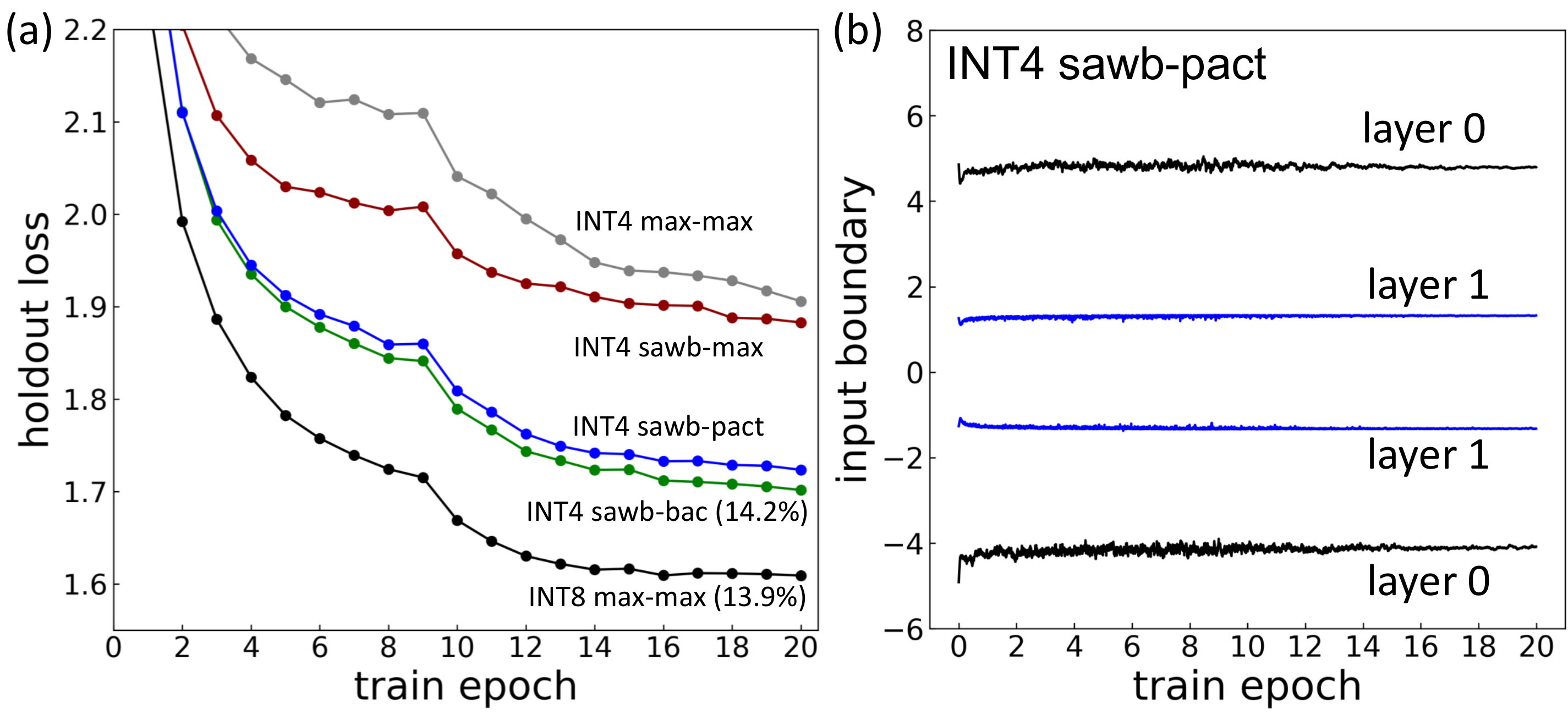}
  \caption{(a) INT quantization of weights and activations on DBLSTM-HMM-300 with different quantizers (avg WER in brackets). (b) Input boundaries $\alpha$ of the first 2 LSTM layers}
  \label{fig:SWB300-quantizers}
\vspace{-4mm}
\end{figure}

\subsection{QAT fine-tuning} \label{ssec:fine-tuning}
INT4 QAT of large speech LSTM models requires extensive computational resources, which hinders its application in many practical scenarios. With the aim of shortening the QAT time, we initialize our networks with state-of-the-art FP32 pre-trained models. We then apply a decreasing LR policy, which gently adapts the pre-trained weights to compensate for the noise introduced by the quantization process. Specifically, in DBLSTM-HMMs we modify the LR schedule into a linear decay (by a 0.5 factor), replacing the constant LR or its warm-up. For RNN-T, we customize the LR schedule to start from a small initial value ($4\cdot10^{-4}$), decrease it to $10^{-5}$ over 8 epochs, and stabilize it afterwards (Fig.~\ref{fig:finetuning}(d)). We observed that our QAT fine-tuning strategy significantly reduces INT4 QAT training time, thus enabling optimization of large LSTM-based speech models (see details in the next section).

\section{Experimental results} \label{sec:experimental}

\sisetup{detect-weight=true,detect-inline-weight=math}
\begin{table}[t]
  \centering
  \caption{4-bit DBLSTM-HMM quantization. $^a$10-epoch fine-tuning from pre-trained FP32 model}
  \label{tab:DBLSTM-HMM}
  \scalebox{0.92}{
    \begin{tabular}{@{} l| ll| c| S[table-format=3.2] @{} S[table-format=3.2] @{} S[table-format=3.2] @{}}
      \toprule
      \multicolumn{1}{c|}{\textbf{model}} & \multicolumn{2}{c|}{\textbf{quantizer}} & \multicolumn{1}{c|}{\textbf{bits}}  & \multicolumn{3}{c}{\textbf{WER}} \\
      \midrule
      \multicolumn{1}{c|}{\textbf{}}   &
      \multicolumn{1}{c}{\textbf{W}}   & \multicolumn{1}{c|}{\textbf{A}} & \multicolumn{1}{c|}{\textbf{W\&A}} & \multicolumn{1}{c}{\textbf{SWB}} & \multicolumn{1}{c}{\textbf{CH}} & \multicolumn{1}{c}{\textbf{avg}} \\
      \midrule
      HMM-300      & -    & -    & 32 &  9.9 & 17.9 & 13.9 \\
      HMM-300      & max  & max  & 8  & 10.0 & 17.7 & 13.9 \\
      HMM-300      & max  & max  & 4  & 11.0 & 19.0 & 15.0 \\
      HMM-300      & sawb & max  & 4  & 10.7 & 18.5 & 14.6 \\
      HMM-300      & sawb & pact & 4  & 10.5 & 18.2 & 14.4 \\
      HMM-300      & sawb & bac  & 4  & 10.2 & 18.2 & 14.2 \\
      \bfseries HMM-300$^a$  & \bfseries sawb & \bfseries bac  & \bfseries 4  & \bfseries 10.4 & \bfseries 18.3 & \bfseries 14.4 \\
      \midrule    
      HMM-2000     & -    & -    & 32 & 7.6  & 13.0 & 10.3 \\
      HMM-2000     & max  & max  & 8  & 7.5  & 13.0 & 10.3 \\
      HMM-2000     & max  & max  & 4  & 9.8  & 16.2 & 13.0 \\
      HMM-2000     & sawb & max  & 4  & 9.0  & 15.6 & 12.3 \\
      HMM-2000     & sawb & pact & 4  & 8.1  & 14.2 & 11.2 \\
      HMM-2000     & sawb & bac  & 4  & 8.0  & 14.2 & 11.1 \\
      \bfseries HMM-2000$^a$ & \bfseries sawb & \bfseries bac  & \bfseries 4  & \bfseries 8.0  & \bfseries 14.0 & \bfseries 11.0
    \end{tabular}
  }
\vspace{-6mm}
\end{table}

\subsection{DBLSTM-HMM} \label{ssec:hmm}

Figure~\ref{fig:SWB300-quantizers}(a) presents the evolution of holdout losses of the DBLSTM-HMM-300 model during a 20-epoch training for different QAT quantizers. INT8 quantization of weights and activations with the simplest MAX quantizer closely matches the FP32 results in both losses and WER, while the INT4 MAX quantizer shows degraded performance in terms of losses and WER, for a SWB / CH WER gap with the FP32 baseline of 1.1\% / 1.1\%, as shown in Table~\ref{tab:DBLSTM-HMM}. Replacing the MAX quantizer for the LSTM and FC \textit{weights} with SAWB produces some improvements in both losses and WER. Although the gain in losses is limited ($<$2\% in relative terms), the improvement in WER is more pronounced, shrinking the SWB / CH accuracy gap to 0.8\% / 0.6\%. Further improvements are obtained by quantizing activations (inputs and hidden states) with asymmetric PACT: loss degradation is significantly reduced and the average WER improves 0.2\% compared to the SAWB/MAX combination.

Figure~\ref{fig:SWB300-quantizers}(b) shows the evolution of the boundaries $\alpha_{\pm}$ for the \textit{inputs} of the first two LSTM layers with a PACT quantizer. For layers beyond the first, these values converge within a single epoch to the constraints set by the LSTM cell ($\pm$1), scaled by the dropout layer to $\pm1.3$. A similar behavior is seen for the boundaries on the hidden states, which quickly approach $\pm1$ (no dropout scaling is applied in between time steps). In separate experiments, we also observed that the convergence values of the boundaries in the first layer, which do take several epochs to settle, are not very sensitive to their initial choice.

The BAC quantizer obtains the best results when applied to the LSTM layers of DBLSTM-HMM-300, with a slight improvement in losses compared to PACT and a WER gap of just 0.3\% on either test set, compared to the FP32 model. Training time is $\sim$10\% faster than PACT. Qualitatively, the learned boundaries $\alpha_{\pm}$ for the first layer follow a similar trend as with PACT, and are constant in all following layers (see Sec.~\ref{ssec:quantizers}). DBLSTM-HMM-2000 QAT displays a similar behavior: at 8 bits, a simple MAX quantizer is sufficient to preserve the accuracy of the FP32 model (WER SWB/CH = 7.5/13.0\%). On the other hand, INT4 QAT with SAWB for weights and PACT for activations shows 1\% degradation compared to the best performing model. Replacing PACT with BAC results in our best average WER (0.8\% gap), with a reduction in training time of $\sim$10\%.

\begin{figure}[t]
  \centering
  \includegraphics[width=\linewidth]{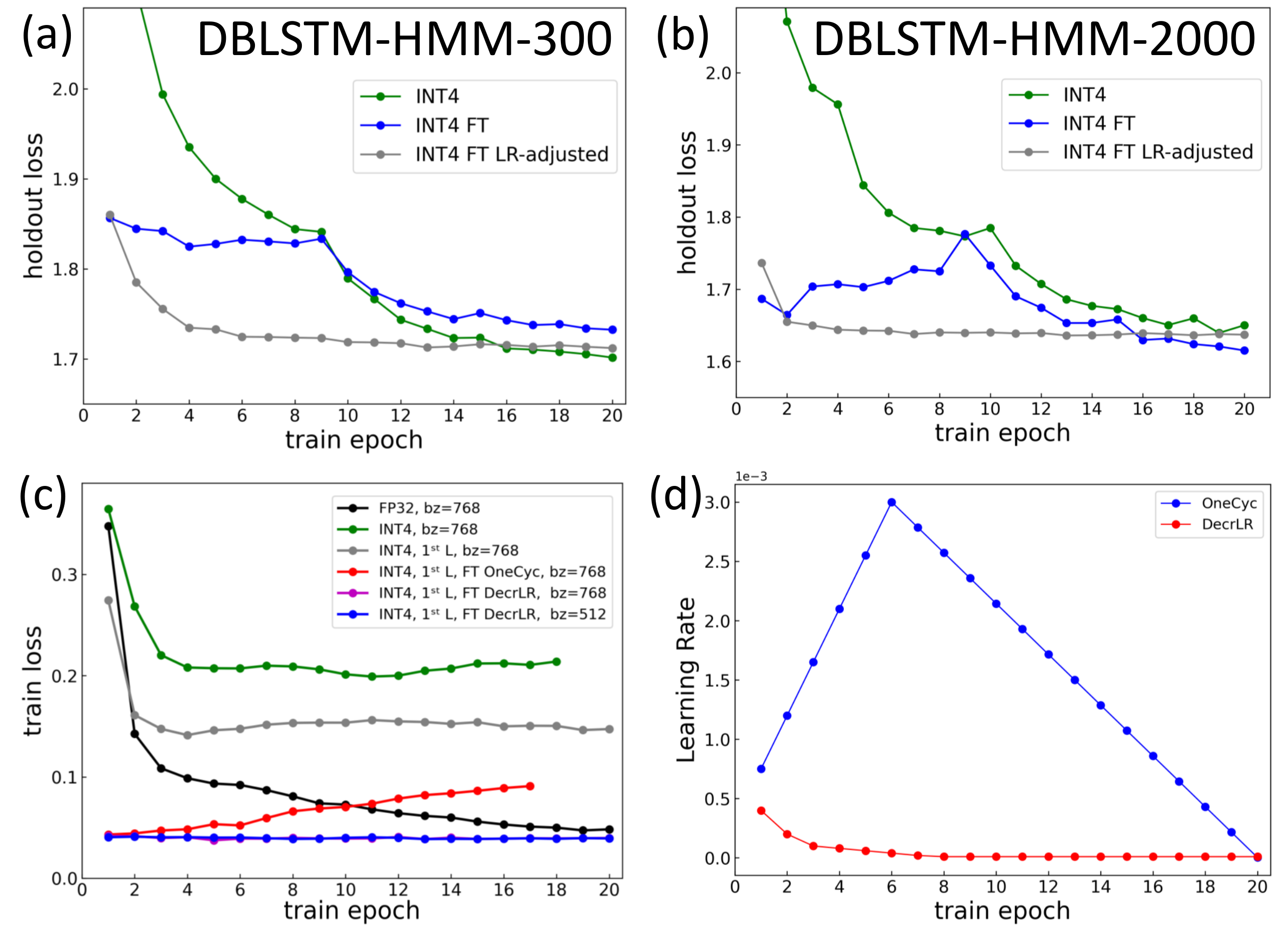}
  \caption{(a,b) DBLSTM-HMM QAT fine-tuning (FT) from pre-trained model. (c) RNN-T QAT, with and without fine-tuning. (d) RNN-T QAT LR schedules (triangular and decreasing)}
  \label{fig:finetuning}
\vspace{-5mm}
\end{figure}

In addition of adopting the BAC quantizer, we apply our QAT fine-tuning strategy to achieve state-of-art INT4 DBLSTM-HMM models. Figure~\ref{fig:finetuning}(a) compares the losses during QAT of DBLSTM-HMM-300 trained either from scratch or from a pre-trained FP32 model. By adjusting the LR schedule as described in Sec.~\ref{ssec:fine-tuning}, training time is halved: by epoch 10 losses have stabilized and a WER of 10.4/18.3\% is obtained. As shown in Table~\ref{tab:DBLSTM-HMM}, with significantly longer training time, QAT from scratch may achieve marginally better results. This strategy however does not scale well in larger LSTM-based models such as DBLSTM-HMM-2000 and RNN-T. Figure~\ref{fig:finetuning}(b) compares losses in similar scenarios using the DBLSTM-HMM-2000 model. The QAT fine-tuning strategy rapidly achieves small holdout losses within 5 epoch and results in the best WER (SWB/CH = 8.0/14.0\%), outperforming QAT from scratch with only half of the training time.

\subsection{RNN-T}
\label{ssec:rnnt}

RNN-T training is computationally expensive, which makes QAT much harder than for other workloads~\cite{li2019improving}. To speed up training, we adopt large-batch training techniques~\cite{goyal2017accurate} and use 24 V100 GPUs with a per-worker batch size of 32. We also leverage BAC, with $\alpha = 1.25$ on the inputs and $\alpha = 1$ on the hidden states. Compared to QAT with learnable quantizer parameters (PACT), BAC achieves a 1.4$\times$ speed-up in this RNN-T implementation. Although these solutions make INT4 QAT exploration more efficient, they are not sufficient to achieve high accuracy. As shown in Fig.~\ref{fig:finetuning}(c) (green line), INT4 QAT causes a significant degradation in losses compared to the FP32 baseline (black line). To improve accuracy we further apply our QAT fine tuning strategy (see Sec.~\ref{ssec:fine-tuning}) in RNN-T. In detail, (1) since a proper representation of the first LSTM layer encoder inputs is critical and the impact of this layer on computation is limited (see  Sec.~\ref{ssec:models_quant}), we retain it in full-precision. This decisively mitigates the degradation (grey line). (2) We apply QAT to fine-tune a state-of-the-art pre-trained model (red line)\footnote{QAT fine tuning from FP32 model with batch size (64) shows comparable training losses to FP32 training with large batch size (768)}; it has been shown that starting from a pre-trained model requires significantly fewer train iterations~\cite{krishnamoorthi2018quantizing}; this is particularly important for RNN-T. (3) Instead of OneCycleLR~\cite{saon2021advancing}, we use a custom LR policy (see Sec.~\ref{ssec:fine-tuning} and Fig.~\ref{fig:finetuning}(d)) which achieves better performance (purple line). (4) To further improve the QAT result, we reduce the total batch size from 768 to 512; this mitigates large batch size training challenges~\cite{you2019large} with an acceptable trade-off in training time. With help from these tricks, our INT4 QAT quickly reaches a small training loss in 8 epochs and achieves 13.1\% average WER (8.8\% for SWB and 17.4\% for CH), a 1.3\% degradation compared to the FP32 baseline of 11.8\% (WER SWB 8.0\%, CH 15.6\%)~\cite{saon2021advancing}. Table~\ref{tab:RNNT} summarizes our RNN-T quantization experiments.\footnote{In Table~\ref{tab:RNNT}, acronyms are as follows. BL: baseline; FT: fine tuning; BZ: batch size; LR: learning rate; 1\textsuperscript{st} L: 1\textsuperscript{st} LSTM layer of encoder; PrNN: prediction network; OneCyc: one cycle learning rate policy; DecrLR: decreasing learning rate policy. The best WER obtained in QAT is reported.} Notice that a FP32 \textit{prediction} network achieves the best WER (13.05\%) but also increases decoding inference time by 3.3$\times$, compared to a 4-bit prediction network (see Sec.~\ref{ssec:runtime}). Thus, optimizing both accuracy and inference speed requires its quantization.

\sisetup{detect-weight=true,detect-inline-weight=math}
\begin{table}[t]
  \centering
  \caption{4-bit RNN-T quantization.}
  \label{tab:RNNT}
  \scalebox{0.84}{
  \begin{tabular}{@{} c@{} c@{} c@{} c | c @{} c | S[table-format=3.2] @{} S[table-format=3.2] @{} S[table-format=3.2] @{}}
    \toprule
    \multicolumn{1}{c}{\textbf{QAT}} & \multicolumn{1}{c}{\textbf{FT}} &
    \multicolumn{1}{c}{\textbf{BZ}}  & \multicolumn{1}{c|}{\textbf{LR}} &
    \multicolumn{2}{c|}{\textbf{bits}} & \multicolumn{3}{c}{\textbf{WER}} \\
    \midrule
    \multicolumn{4}{c|}{\textbf{}}  & \multicolumn{1}{c}{\textbf{1\textsuperscript{st}~L}} 
    & \multicolumn{1}{c|}{\textbf{PrNN}} & \multicolumn{1}{c}{\textbf{SWB}} &
    \multicolumn{1}{c}{\textbf{CH}} & \multicolumn{1}{c}{\textbf{avg}} \\
    \midrule
    BL  & BL   & 64 & OneCyc  & 32 & 32 & 8.0  & 15.6 & 11.8  \\
    N   & --  & --  & --      & 4  & 4 & 12.2 & 22.6  & 17.4  \\
    Y   & N   & 768 & OneCyc  & 4  & 4 & 30.1 & 40.9  & 35.5  \\
    Y   & N   & 768 & OneCyc  & 32 & 4 & 17.3 & 27.3  & 22.3  \\
    Y   & Y   & 768 & OneCyc  & 32 & 4 & 9.8  & 18.7  & 14.25 \\
    Y   & Y   & 768 & DecrLR  & 32 & 4 & 9.2  & 17.4  & 13.3  \\
   \bfseries Y & \bfseries Y   & \bfseries 512 & \bfseries DecrLR  & \bfseries 32 & \bfseries 4 & \bfseries 8.8 & \bfseries 17.4 & \bfseries 13.1 \\
    Y   & Y   & 512 & DecrLR  & 32 & 32 & 8.8 & 17.3 & 13.05
  \end{tabular}}
\vspace{-4mm}
\end{table}

\subsection{Inference performance in end-to-end models}
\label{ssec:runtime}

We quantify the runtime improvement achieved through INT4 quantization relative to FP16 implementation. We consider a system with a CPU attached to a co-processor capable of executing common deep learning operations such as GEMM, LSTM activation \emph{etc.} at various data precisions~\cite{Agrawal2021}. The co-processor executes the compute-heavy data-parallel portions of the workload, while control-heavy operations (\emph{e.g.}, sorting and pruning hypothesis) are mapped to the CPU. We utilize a detailed performance model~\cite{Venkataramani2019,venkataramani2019memory} to estimate runtime of the co-processor, which has been calibrated to within 1\% of silicon measurements. We further instrumented our CPU implementation with performance counters to measure the execution time of workload portions mapped to the CPU.

Fig.~\ref{fig:problem-intro}(a) shows the improvement in INT4 inference runtime on the RNN-T model for an \emph{average} sequence length of 152 and different beam widths. The total execution time is broken into the encoder, prediction network, and joint components. In the acoustic encoder, all LSTM layers beside the first are quantized to 4 bits; this dramatically increases their computation throughput and reduces encoder runtime by 2.6$\times$ (blue bars). Due to the iterative beam search (decoding) process~\cite{saon2021advancing}, the decoder runtime increases significantly with beam width. Thanks to the quantized prediction network, the decoding time (red bars) scales well between FP16 and INT4, achieving 3.3$\times$ speed-up, mitigating the impact of wider beams. On the other hand, the joint network runtime (gray bars) remains constant, as it is always executed in FP16 and involves control-heavy operations executed on the CPU. As the fraction of operations contributed by the joint network is small, this has little impact on the end-to-end runtime. Overall, the INT4 implementation achieves 2.6$\times$ speed-up compared to FP16 across different beam widths.\footnote{As beam width increases, so does the proportion of both the prediction network (red) and non-accelerable computations (gray), resulting in the beam width having minor impact on the end-to-end acceleration.}

\section{Conclusions}

4-bit quantization is a promising technique to dramatically reduce inference time in LSTM-based ASR systems, but has not yet found wide adoption in real-world scenarios, primarily due to lack of demonstrations on large models and standard datasets, and concerns about accuracy degradation. In this paper, we proposed a practical methodology to achieve state-of-art accuracy at 4 bits, in hybrid and end-to-end LSTM speech networks trained on large corpora and evaluated on a standard test set. Our work lays down the foundation to introduce ultra-low precision inference in speech recognition domains.

\bibliographystyle{IEEEtran}

\bibliography{references}

\begin{thebibliography}{10}
\providecommand{\url}[1]{#1}
\csname url@samestyle\endcsname
\providecommand{\newblock}{\relax}
\providecommand{\bibinfo}[2]{#2}
\providecommand{\BIBentrySTDinterwordspacing}{\spaceskip=0pt\relax}
\providecommand{\BIBentryALTinterwordstretchfactor}{4}
\providecommand{\BIBentryALTinterwordspacing}{\spaceskip=\fontdimen2\font plus
\BIBentryALTinterwordstretchfactor\fontdimen3\font minus
  \fontdimen4\font\relax}
\providecommand{\BIBforeignlanguage}[2]{{%
\expandafter\ifx\csname l@#1\endcsname\relax
\typeout{** WARNING: IEEEtran.bst: No hyphenation pattern has been}%
\typeout{** loaded for the language `#1'. Using the pattern for}%
\typeout{** the default language instead.}%
\else
\language=\csname l@#1\endcsname
\fi
#2}}
\providecommand{\BIBdecl}{\relax}
\BIBdecl

\bibitem{saon2017english}
G.~Saon, G.~Kurata, T.~Sercu, K.~Audhkhasi, S.~Thomas, D.~Dimitriadis, X.~Cui,
  B.~Ramabhadran, M.~Picheny, L.-L. Lim, B.~Roomi, and P.~Hall, ``English
  conversational telephone speech recognition by humans and machines,'' in
  \emph{Proc. Interspeech}, 2017, pp. 132--136.

\bibitem{tuske2020single}
Z.~T{\"u}ske, G.~Saon, K.~Audhkhasi, and B.~Kingsbury, ``Single headed
  attention based sequence-to-sequence model for state-of-the-art results on
  switchboard-300,'' in \emph{Proc. Interspeech}, 2020, pp. 551--555.

\bibitem{saon2021advancing}
G.~Saon, Z.~T{\"u}ske, D.~Bolanos, and B.~Kingsbury, ``Advancing rnn transducer
  technology for speech recognition,'' \emph{arXiv preprint arXiv:2103.09935},
  2021.

\bibitem{godfrey1992switchboard}
J.~J. Godfrey, E.~C. Holliman, and J.~McDaniel, ``Switchboard: Telephone speech
  corpus for research and development,'' in \emph{1992 IEEE International
  Conference on Acoustics, Speech and Signal Processing (ICASSP)}, vol.~1,
  1992, pp. 517--520.

\bibitem{wong2016sequence}
J.~H. Wong and M.~J. Gales, ``Sequence student-teacher training of deep neural
  networks,'' in \emph{Proc. Interspeech}, 2016, pp. 2761--2765.

\bibitem{chebotar2016distilling}
Y.~Chebotar and A.~Waters, ``Distilling knowledge from ensembles of neural
  networks for speech recognition.'' in \emph{Proc. Interspeech}, 2016, pp.
  3439--3443.

\bibitem{sainath2013low}
T.~N. Sainath, B.~Kingsbury, V.~Sindhwani, E.~Arisoy, and B.~Ramabhadran,
  ``Low-rank matrix factorization for deep neural network training with
  high-dimensional output targets,'' in \emph{2013 IEEE International
  Conference on Acoustics, Speech and Signal Processing (ICASSP)}, 2013, pp.
  6655--6659.

\bibitem{seide2011conversational}
F.~Seide, G.~Li, and D.~Yu, ``Conversational speech transcription using
  context-dependent deep neural networks,'' in \emph{Proc. Interspeech}, 2011,
  pp. 437--440.

\bibitem{hubara2017quantized}
I.~Hubara, M.~Courbariaux, D.~Soudry, R.~El-Yaniv, and Y.~Bengio, ``Quantized
  neural networks: Training neural networks with low precision weights and
  activations,'' \emph{The Journal of Machine Learning Research}, vol.~18,
  no.~1, pp. 6869--6898, 2017.

\bibitem{he2019streaming}
Y.~He, T.~N. Sainath, R.~Prabhavalkar, I.~McGraw, R.~Alvarez, D.~Zhao,
  D.~Rybach, A.~Kannan, Y.~Wu, R.~Pang \emph{et~al.}, ``Streaming end-to-end
  speech recognition for mobile devices,'' in \emph{2019 IEEE International
  Conference on Acoustics, Speech and Signal Processing (ICASSP)}, 2019, pp.
  6381--6385.

\bibitem{shangguan2019optimizing}
Y.~Shangguan, J.~Li, Q.~Liang, R.~Alvarez, and I.~McGraw, ``Optimizing speech
  recognition for the edge,'' \emph{arXiv preprint arXiv:1909.12408}, 2019.

\bibitem{wu2020integer}
H.~Wu, P.~Judd, X.~Zhang, M.~Isaev, and P.~Micikevicius, ``Integer quantization
  for deep learning inference: Principles and empirical evaluation,''
  \emph{arXiv preprint arXiv:2004.09602}, 2020.

\bibitem{nguyen2020quantization}
H.~D. Nguyen, A.~Alexandridis, and A.~Mouchtaris, ``Quantization aware training
  with absolute-cosine regularization for automatic speech recognition,'' in
  \emph{Proc. Interspeech}, 2020, pp. 3366--3370.

\bibitem{ardakani2018learning}
A.~Ardakani, Z.~Ji, S.~C. Smithson, B.~H. Meyer, and W.~J. Gross, ``Learning
  recurrent binary/ternary weights,'' in \emph{International Conference on
  Learning Representations}, 2019.

\bibitem{hou2019normalization}
L.~Hou, J.~Zhu, J.~Kwok, F.~Gao, T.~Qin, and T.-Y. Liu, ``Normalization helps
  training of quantized lstm,'' in \emph{Advances in Neural Information
  Processing Systems}, vol.~32, 2019.

\bibitem{he2016effective}
Q.~He, H.~Wen, S.~Zhou, Y.~Wu, C.~Yao, X.~Zhou, and Y.~Zou, ``Effective
  quantization methods for recurrent neural networks,'' \emph{arXiv preprint
  arXiv:1611.10176}, 2016.

\bibitem{xu2018alternating}
C.~Xu, J.~Yao, Z.~Lin, W.~Ou, Y.~Cao, Z.~Wang, and H.~Zha, ``Alternating
  multi-bit quantization for recurrent neural networks,'' in
  \emph{International Conference on Learning Representations}, 2018.

\bibitem{xu2020low}
J.~Xu, X.~Chen, S.~Hu, J.~Yu, X.~Liu, and H.~Meng, ``Low-bit quantization of
  recurrent neural network language models using alternating direction methods
  of multipliers,'' in \emph{2020 IEEE International Conference on Acoustics,
  Speech and Signal Processing (ICASSP)}, 2020, pp. 7939--7943.

\bibitem{cui2017embedding}
X.~Cui, V.~Goel, and G.~Saon, ``Embedding-based speaker adaptive training of
  deep neural networks,'' in \emph{Proc. Interspeech}, 2017, pp. 122--126.

\bibitem{zhang2019highly}
W.~Zhang, X.~Cui, U.~Finkler, G.~Saon, A.~Kayi, A.~Buyuktosunoglu,
  B.~Kingsbury, D.~Kung, and M.~Picheny, ``A highly efficient distributed deep
  learning system for automatic speech recognition,'' in \emph{2019 IEEE
  International Conference on Acoustics, Speech and Signal Processing
  (ICASSP)}, 2019, pp. 5706--5710.

\bibitem{ko2015audio}
T.~Ko, V.~Peddinti, D.~Povey, and S.~Khudanpur, ``Audio augmentation for speech
  recognition,'' in \emph{2015 IEEE International Conference on Acoustics,
  Speech and Signal Processing (ICASSP)}, 2015, pp. 3586--3589.

\bibitem{park2019specaugment}
D.~S. Park, W.~Chan, Y.~Zhang, C.-c. Chiu, B.~Zoph, E.~D. Cubuk, and Q.~V. Le,
  ``{SpecAugment}: A simple data augmentation method for automatic speech
  recognition,'' in \emph{Proc. Interspeech}, 2019, pp. 2613--2617.

\bibitem{saon2019sequence}
G.~Saon, Z.~T{\"u}ske, K.~Audhkhasi, and B.~Kingsbury, ``Sequence noise
  injected training for end-to-end speech recognition,'' in \emph{2019 IEEE
  International Conference on Acoustics, Speech and Signal Processing
  (ICASSP)}, 2019, pp. 6261--6265.

\bibitem{choi2018bridging}
J.~Choi, P.~I.-J. Chuang, Z.~Wang, S.~Venkataramani, V.~Srinivasan, and
  K.~Gopalakrishnan, ``Bridging the accuracy gap for 2-bit quantized neural
  networks (qnn),'' \emph{arXiv preprint arXiv:1807.06964}, 2018.

\bibitem{bengio2013estimating}
Y.~Bengio, N.~L{\'e}onard, and A.~Courville, ``Estimating or propagating
  gradients through stochastic neurons for conditional computation,''
  \emph{arXiv preprint arXiv:1308.3432}, 2013.

\bibitem{li2019improving}
J.~Li, R.~Zhao, H.~Hu, and Y.~Gong, ``Improving rnn transducer modeling for
  end-to-end speech recognition,'' in \emph{2019 IEEE Automatic Speech
  Recognition and Understanding Workshop (ASRU)}.\hskip 1em plus 0.5em minus
  0.4em\relax IEEE, 2019, pp. 114--121.

\bibitem{goyal2017accurate}
P.~Goyal, P.~Doll{\'a}r, R.~Girshick, P.~Noordhuis, L.~Wesolowski, A.~Kyrola,
  A.~Tulloch, Y.~Jia, and K.~He, ``Accurate, large minibatch sgd: Training
  imagenet in 1 hour,'' \emph{arXiv preprint arXiv:1706.02677}, 2017.

\bibitem{krishnamoorthi2018quantizing}
R.~Krishnamoorthi, ``Quantizing deep convolutional networks for efficient
  inference: A whitepaper,'' \emph{arXiv preprint arXiv:1806.08342}, 2018.

\bibitem{you2019large}
Y.~You, J.~Hseu, C.~Ying, J.~Demmel, K.~Keutzer, and C.-J. Hsieh, ``Large-batch
  training for lstm and beyond,'' in \emph{Proceedings of the International
  Conference for High Performance Computing, Networking, Storage and Analysis},
  2019, pp. 1--16.

\bibitem{Agrawal2021}
A.~{Agrawal}, S.~K. {Lee}, J.~{Silberman}, M.~{Ziegler}, M.~{Kang},
  S.~{Venkataramani}, N.~{Cao}, B.~{Fleischer}, M.~{Guillorn}, M.~{Cohen},
  S.~{Mueller}, J.~{Oh}, M.~{Lutz}, J.~{Jung}, S.~{Koswatta}, C.~{Zhou},
  V.~{Zalani}, J.~{Bonanno}, R.~{Casatuta}, C.~Y. {Chen}, J.~{Choi},
  H.~{Haynie}, A.~{Herbert}, R.~{Jain}, M.~{Kar}, K.~H. {Kim}, Y.~{Li},
  Z.~{Ren}, S.~{Rider}, M.~{Schaal}, K.~{Schelm}, M.~{Scheuermann}, X.~{Sun},
  H.~{Tran}, N.~{Wang}, W.~{Wang}, X.~{Zhang}, V.~{Shah}, B.~{Curran},
  V.~{Srinivasan}, P.~F. {Lu}, S.~{Shukla}, L.~{Chang}, and
  K.~{Gopalakrishnan}, ``9.1 a 7~nm 4-core ai chip with 25.6~tflops hybrid fp8
  training, 102.4~tops int4 inference and workload-aware throttling,'' in
  \emph{2021 IEEE International Solid- State Circuits Conference (ISSCC)},
  vol.~64, 2021, pp. 144--146.

\bibitem{Venkataramani2019}
S.~{Venkataramani}, J.~{Choi}, V.~{Srinivasan}, W.~{Wang}, J.~{Zhang},
  M.~{Schaal}, M.~J. {Serrano}, K.~{Ishizaki}, H.~{Inoue}, E.~{Ogawa},
  M.~{Ohara}, L.~{Chang}, and K.~{Gopalakrishnan}, ``Deeptools: Compiler and
  execution runtime extensions for rapid ai accelerator,'' \emph{IEEE Micro},
  vol.~39, no.~5, pp. 102--111, 2019.

\bibitem{venkataramani2019memory}
S.~Venkataramani, V.~Srinivasan, J.~Choi, P.~Heidelberger, L.~Chang, and
  K.~Gopalakrishnan, ``Memory and interconnect optimizations for peta-scale
  deep learning systems,'' in \emph{2019 IEEE 26th International Conference on
  High Performance Computing, Data, and Analytics (HiPC)}, 2019, pp. 225--234.

\end{thebibliography}

\end{document}